\documentclass{article}

\usepackage{arxiv}

\usepackage[utf8]{inputenc} 
\usepackage[T1]{fontenc}    
\usepackage{hyperref}       
\usepackage{url}            
\usepackage{booktabs}       
\usepackage{amsfonts}       
\usepackage{nicefrac}       
\usepackage{microtype}      
\usepackage{lipsum}
\usepackage{graphicx}
\graphicspath{ {./images/} }

\usepackage{cite}
\usepackage{amsmath,amssymb}
\usepackage{algorithmic}
\usepackage{caption}
\usepackage{enumerate}
\usepackage{enumitem}
\usepackage{hhline, booktabs}

\usepackage{graphicx}
\usepackage{textcomp}
\usepackage{multirow}

\title{A Laplacian-based Quantum Graph Neural Network for Semi-Supervised Learning}

\author{
 Hamed Gholipour\\
 Department of Computer Science, University of Beira Interior, Covilhã, 6200-001, Portugal \\
  \texttt{hamed.gholipour@ubi.pt} \\
   \And
 Farid Bozorgnia \\
  Department of Mathematics, 
  Instituto Superior Técnico,
  Lisbon, 1049-001 Portugal \\
  \texttt{bozorgnia@math.ist.utl.pt} \\
  \And
  Kailash Hambarde \\
  Department of Computer Science,
  University of Beira Interior,
  Covilhã, 6201-001 Portugal \\
  \texttt{kailas.srt@gmail.com} \\
  \And
 Hamzeh Mohammadigheymasi \\
 Department of Computer Science, University of Beira Interior, Covilhã, 6200-001, Portugal\\
 Atmosphere and Ocean Research Institute, The University of Tokyo, Kashiwa, 277-0882, Japan   \\
  \texttt{hamzeh@ubi.pt} \\
\And
   Javier Mancilla \\
  Falcondale LLC,  Vigo, Spain \\
  \texttt{javier@falcondale.pro} \\
  \And
 Andre Sequeira \\
  Department of Informatics,
  INESC TEC, University of Minho,
  Braga, 4710057 Portugal \\
  \texttt{andre.m.sequeira@inesctec.pt} \\
\And
   Joao Neves\\
Department of Computer Science,
 University of Beira Interior,
  Covilhã, 6201-001 Portugal \\
  \texttt{jcneves@ubi.pt} \\
  \And
 Hugo Proença \\
 Department of Computer Science,
 University of Beira Interior,
 Covilhã, 6201-001 Portugal \\
 \texttt{hugomcp@di.ubi.pt} \\
}

\begin{document}

\maketitle
\newpage
\begin{abstract}
Laplacian learning method is a well-established technique in classical graph-based semi-supervised learning, but its potential in the quantum domain remains largely unexplored. This study investigates the performance of the Laplacian-based Quantum Semi-Supervised Learning (QSSL) method across four benchmark datasets—Iris, Wine, Breast Cancer Wisconsin, and Heart Disease. Further analysis explores the impact of increasing qubit counts, revealing that adding more qubits to a quantum system doesn't always improve performance. The effectiveness of additional qubits depends on the quantum algorithm and how well it matches the dataset. 
Additionally, we examine the effects of varying entangling layers on entanglement entropy and test accuracy. The performance of Laplacian learning is highly dependent on the number of entangling layers, with optimal configurations varying across different datasets. Typically, moderate levels of entanglement offer the best balance between model complexity and generalization capabilities. These observations highlight the crucial need for precise hyperparameter tuning tailored to each dataset to achieve optimal performance in Laplacian learning methods.
\end{abstract}


\section{Introduction}
\label{sec:introduction}
\subsection{related work}
The development of machine learning techniques has undergone significant transformations over the past few decades, evolving from simple linear models to sophisticated deep learning architectures \cite{mahesh2020machine}. Initially, machine learning methods focused on supervised learning, where models were trained on fully labeled datasets to predict outcomes on unseen data \cite{zhu2022introduction,mohammadigheymasi2023application}. Supervised Learning (SL) can be efficiently utilized in other fields such as biomedical engineering, finance, and Earth and environmental sciences, providing robust solutions for predictive modelling and data analysis
\cite{10177744}. However, the growing need to handle vast amounts of unlabeled data led to unsupervised learning techniques to find patterns and structures in data without labeled outcomes. Also, this progression laid the groundwork for SSL; the hybrid approach combines the strengths of supervised and unsupervised learning to support labeled and unlabeled data, thereby enhancing model accuracy and robustness \cite{9700632}. SSL is a technique that mitigates the challenge of limited labeled data by utilizing a mix of labeled and unlabeled data during training \cite{zhu2022introduction}. SSL bridges the gap between supervised and unsupervised learning, leveraging vast amounts of unlabeled data to improve model performance.
Integrating quantum computing with machine learning represents the latest frontier in this evolutionary journey \cite{10085586}. One of the most innovative advancements in Quantum Machine Learning (QML) is Quantum Graph Learning, which synergizes quantum computing with graph-based learning methodologies \cite{10391361}. By incorporating quantum circuits into GNNs, QGL offers a powerful framework to address complex challenges in graph learning. This novel approach holds promise for various applications, from optimizing network communications to advancing drug Analysis \cite{mishra2021quantum}. 
Graph-based SSL, representing data points as nodes connected by edges denoting relationships or similarities, propagates label information across the graph to infer labels for unlabeled nodes. This method includes label propagation and graph-based regularization to maintain label consistency among neighbouring nodes. Laplacian is a prominent graph-based SSL method. Laplacian learning, initially detailed by Zhu et al. \cite{zhu2003semi}, has demonstrated strong performance in some scenarios but falls short in others, particularly in classification tasks in a few label \cite{flores2022analysis}. 
\subsection{Contributions}
This study investigates the integration of Quantum Graph Neural Networks within the framework of Lapalacian-based QSSL. It assesses the effectiveness of Laplacian QSSL methods across four benchmark datasets: Iris\cite{misc_iris_53}, Wine \cite{misc_wine_109}, Breast Cancer Wisconsin Diagnostic\cite{misc_breast_cancer_wisconsin_(diagnostic)_17}, and Heart Disease\cite{misc_heart_disease_45}. The upcoming evaluation will systematically investigate the effects of varying qubit counts and the number of entangling layers on test accuracy.  The observation that increasing the number of qubits does not necessarily enhance performance underscores a complex interplay of factors. Higher qubit counts can result in more intricate quantum circuits and increased error rates, potentially leading to a decline in performance. Increasing qubits generates a barren plateau phenomenon, which leads to the gradient of vanishing cost function exponentially with the number of qubits. So, when the number of qubits increases, it is a significant challenge for scalable quantum computing. As shown, after 14 qubits, the cost function degrades suddenly.
Moreover, quantum algorithms like the Laplacian method may not be optimized for larger qubit configurations, leading to a disproportionate increase in computational burden. The effectiveness of additional qubits also depends on the dataset's characteristics, as they may not always align well with the data's features. Optimising quantum algorithms, managing resources efficiently, and carefully tuning models to balance complexity and performance are essential to address these challenges. Understanding these dynamics is crucial for advancing quantum machine learning and fully leveraging the benefits of quantum computing. 

The performance of the Laplacian learning method is highly sensitive to the number of entangling layers, with optimal configurations differing across datasets. Generally, moderate levels of entanglement strike the best balance between model complexity and generalization capability. These findings highlight the importance of hyperparameter tuning specific to each dataset to achieve optimal performance in Laplacian learning methods.

The paper is structured as follows: We begin with an overview of Quantum Graph Learning \cite{xia2021graph} and its relevance to semi-supervised learning. We then delve into the Laplacian QSSL methods, followed by a detailed analysis of their performance across different datasets. Afterwards, we will review our adjustments to the qubit and entangling layers based on test accuracy and entanglement in QSSL methods.

\begin{figure*}[htbp]
  \centering
  \includegraphics[width=0.7\textwidth]{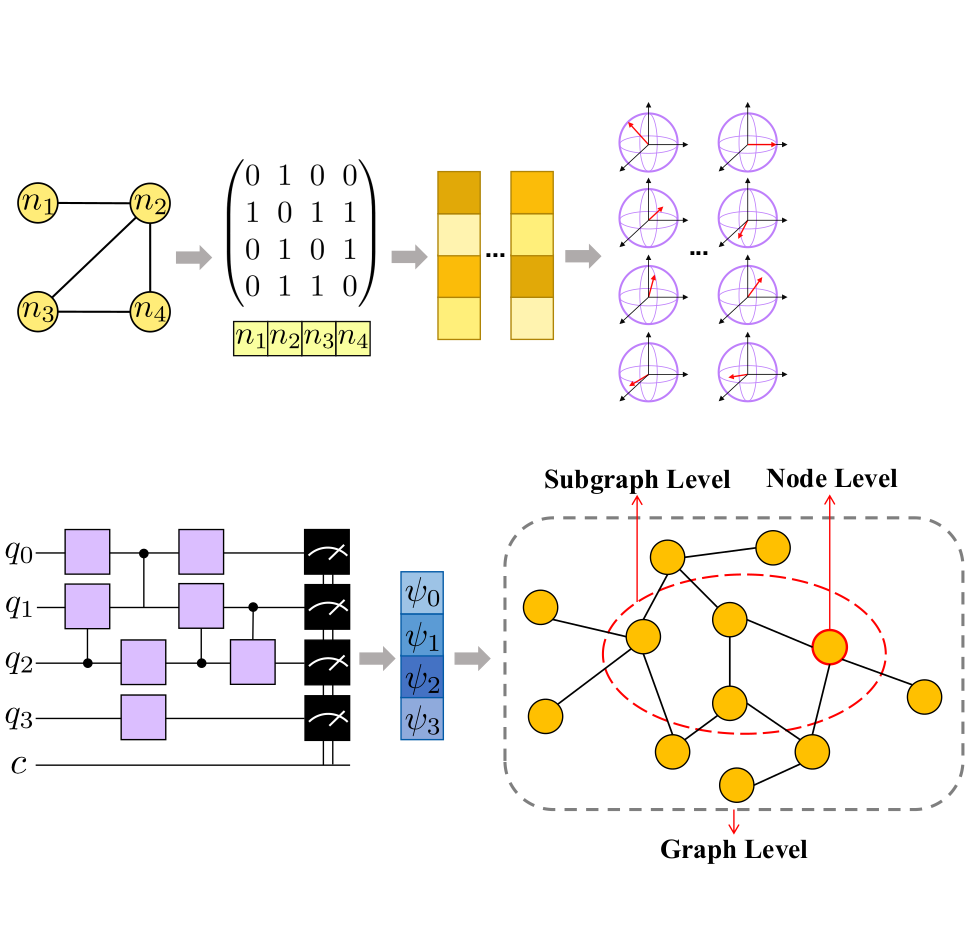}
  \caption{A quantum circuit architecture for GNNs, including input encoding, variational quantum layers, and measurement. This hybrid quantum-classical approach aims to enhance GNN performance in tasks such as node classification and link prediction (Figure modified from 
 \cite{yu2023quantum}).}
  \label{fig:example3}
\end{figure*}

\section{Quantum Graph Learning}
In this section, Graph-Based Learning will be explained to clarify how Graphs, called Quantum Graph Learning, can be used in Quantum semi-supervised learning. Graph learning is a field of machine learning focused on algorithms that analyze and interpret data structured as graphs. In a graph, nodes represent entities, while edges depict the relationships between these entities. Graphs can model complex systems such as social networks, biological interactions, and knowledge bases.

In Graph Learning, a fundamental aspect is understanding graph structures, where nodes denote entities and edges represent their connections. These connections can be directed or undirected, and they may have weights that indicate the strength of the relationships between nodes.
A key tool in graph learning is (GNNs). GNNs extend traditional neural networks to handle graph-structured data by incorporating the graph's connectivity into the learning process through message passing, where nodes share and aggregate information from their neighbours to update their features.

Graph Learning encompasses several core tasks. Node classification involves predicting the labels of nodes based on their features and the graph’s structure. Link prediction aims to forecast the existence of connections between nodes, which is useful for recommending new links in social networks or predicting biological interactions. Graph classification assigns labels to entire graphs, such as determining the toxicity of molecules, while clustering groups nodes based on their similarities and relationships to detect communities within networks\cite{errica2019fair}.
To achieve these tasks, graph learning employs various techniques. Spectral methods use the eigenvalues and eigenvectors of graph matrices, such as the Laplacian matrix, for functions like clustering and dimensionality reduction. Spatial methods, such as Graph Convolutional Networks (GCNs), perform operations directly on the graph's structure. Random walk-based techniques like DeepWalk and node2vec generate node sequences to learn low-dimensional node representations, and attention mechanisms in models like Graph Attention Networks GATs allow the model to focus on the most relevant parts of the graph\cite{bronstein2017geometric}.

Graph learning finds applications in diverse domains. Social networks help analyze user interactions, detect communities, and recommend connections. In biological networks, it aids in understanding protein interactions and modeling gene regulatory networks. Knowledge graphs benefit from improved search engines, recommendation systems, and question-answering capabilities.

Overall, graph learning employs the inherent structure of graph data to address complex problems and achieve advanced performance across various applications, demonstrating its importance in modern machine learning and data science \cite{skuzinski2022book}.  QGL applies the computational power of quantum computing and its theoretical foundations to offer solutions for graph learning tasks. QGL methods are classified into three primary types: quantum computing on graphs, quantum graph representation, and quantum circuits designed for graph neural networks GNNs (figure 1). This paper will investigate Quantum Circuits for graph neural networks in more detail. 
Many researchers have turned their attention to quantum graph neural networks with the advent of noisy intermediate-scale Quantum (NISQ) devices. These networks integrate graph neural networks with quantum modules to enhance existing models \cite{wang2019deep}. 

\begin{figure*}[htbp]
  \centering
  \hspace{-2cm}
  \includegraphics[width=0.9\textwidth]{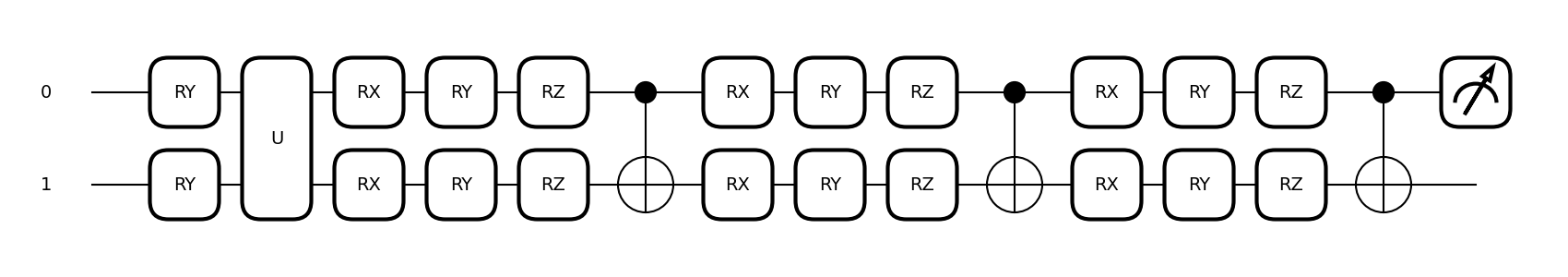}
  \caption{The Quantum Circuit used in the paper, detailing its structure and components including entangling layers, unitary operations, and quantum gates}
  \label{fig:example}
\end{figure*}

An advantage of implementing QGNNs relies on leveraging NISQ devices to adapt the structure of graph neural networks through Parametrized Quantum Circuits(PQSs)\cite{chen2022novel}. NISQ uses parametrised quantum circuits (PQCs), a set of gates with free parameters to be tuned to solve a given task of interest, such as a variational quantum algorithm \cite{cerezo2021variational}. Quantum Graph Neural Network is simulated by PQCs in a way that encodes input graph data in quantum amplitudes(see Figure 1 \cite{yu2023quantum}). The following section will explain how quantum graph learning uses the Laplacian method in QSSL for classification.

\section{Laplacian-based Quantum semi-supervised learning}
SSL techniques serve as a bridge between supervised and unsupervised learning. They leverage extensive unlabeled datasets alongside a smaller set of labeled data to enhance the learning model. SSL aims to extend the known labels to unlabeled samples through a suitable smoothing operator. 

\subsection{Laplacian Learning}

Consider a graph \( G = (V, E) \) with \( n \) nodes. The adjacency matrix \( A \) is an \( n \times n \) matrix where \( A_{ij} \) represents the weight of the edge between nodes \( i \) and \( j \). 

The degree matrix \( D \) is a diagonal matrix where \( D_{ii} = \sum_j A_{ij} \), i.e., the degree of node \( i \). 

The normalized Laplacian matrix \( L_{\text{norm}} \) is defined as:
\[
L_{\text{norm}} = I - D^{-1/2} A D^{-1/2}
\]
where \( I \) is the identity matrix.

Let \( \mathbf{Y} \) be the initial label matrix of size \( n \times c \), where \( n \) is the number of nodes and \( c \) is the number of classes.

To refine the label matrix \( \mathbf{F} \), we use the iterative update rule:
\[
\mathbf{F}^{(t+1)} = \alpha L_{\text{norm}} \mathbf{F}^{(t)} + (1 - \alpha) \mathbf{Y}
\]
Where \( \alpha \) is a weighting factor (typically close to 1) and \( T \) is the number of iterations.

\subsubsection{Iteration Process}

1. Initialize the label matrix:
\[
\mathbf{F}^{(0)} = \mathbf{Y}
\]

2. For each iteration \( t \) from \( 0 \) to \( T-1 \):
\[
\mathbf{F}^{(t+1)} = \alpha L_{\text{norm}} \mathbf{F}^{(t)} + (1 - \alpha) \mathbf{Y}
\]

After \( T \) iterations, the refined label matrix is:
\[
\mathbf{F}^{(T)}
\]

\textbf{Quantum Implementation:}
An adjacency matrix is initially generated from a graph in the quantum realm. Both Poisson and Laplacian matrices can be encoded into the quantum state. This process is referred to as amplitude encoding, where a quantum state \( |\psi\rangle \) can represent the graph data:
\[
|\psi\rangle = \sum_i \alpha_i |i\rangle
\]
with \( \alpha_i \) being the amplitude corresponding to node \( i \).

Solving the Laplacian for a quantum implementation involves quantum algorithms such as the Quantum Phase Estimation (QPE) and the Harrow-Hassidim-Lloyd (HHL) algorithm. These algorithms provide a way to efficiently solve linear systems and eigenvalue problems, which are at the core of Laplacian and Poisson-based methods.

For instance, the HHL algorithm can solve equations of the form \( Ax = b \) efficiently, where \( A \) can be the Laplacian matrix \( L \). The algorithm prepares a quantum state corresponding to \( b \) and then uses quantum operations to find the state corresponding to \( x \).
In practice, the adjacency matrix \( A \), degree matrix \( D \), and consequently the Laplacian matrix \( L \) are encoded into a quantum state. The quantum algorithm then operates on this state to find the desired solution \( f \), which represents the labels' smooth propagation over the graph.
By incorporating these mathematical details and quantum implementation aspects, the paper clarifies how the Laplacian method functions in graph-based semi-supervised learning. This addition ensures that readers can grasp these methods' theoretical underpinnings, mathematical formulation, and potential quantum advantages.
As one of the semi-supervised learning methods, graph-based methods leverage the data's inherent structure, often represented as a graph, to propagate label information from labeled to unlabeled data points. These methods construct a graph where nodes represent data points, and edges capture relationships between them, such as similarity or proximity. By iteratively propagating labels through the Graph, often guided by the Graph's structure or properties, graph-based semi-supervised learning algorithms can effectively generalize known labels to unlabeled samples(\cite{zhu2003semi}). Indeed, a significant portion of graph-based learning methods employs the Graph Laplacian as the smoothing operator to facilitate generalization. However, more sophisticated nonlinear approaches resort to p-Laplacian operators for enhanced performance. These operators enable a more nuanced representation of the data, better capturing complex relationships and structures within the Graph (\cite{elmoataz2017game}). You can find more information about Graph learning in two references (\cite{calder2020poisson}),(\cite{streicher2023graph}).

In the transition to the quantum realm, an adjacency matrix is initially generated from a graph. Subsequently,  Laplacian matrices can be encoded into the quantum state or amplitude of the quantum state vector, a process referred to as Amplitude encoding \cite{setia2018bravyi},\cite{lloyd2010quantum}.

\section{Algorithm}






This algorithm integrates quantum computing with Laplacian learning to perform semi-supervised classification. It utilizes quantum circuits for classification and refines predictions using Laplacian smoothing. Additionally, the algorithm includes parameter tuning and performance evaluation through visualization.

\begin{algorithmic}[1]
\STATE \textbf{Input:} 
\begin{itemize}
    \item Labeled data $(X_l, y_l)$
    \item Unlabeled data $X_u$
    \item Parameters: $\alpha$ (smoothing factor), $T$ (number of Laplacian learning iterations), $K$ (number of optimization iterations), $\eta$ (learning rate), $L$ (number of quantum layers), $n$ (number of qubits)
    \item Hyperparameters: $H_{\text{range}}$ (range for hyperparameter tuning)
\end{itemize}
\STATE \textbf{Output:} 
\begin{itemize}
    \item Predictions for unlabeled data $y_u$
    \item Entropies $S_u$
    \item Performance metrics: Accuracy, Precision, Recall, F1 Score
\end{itemize}

\STATE \textbf{Step 1: Data Preprocessing}
\STATE Fetch the dataset and replace missing values with column medians:
\[
X_{ij} \leftarrow \text{median}(X_{\cdot j}) \text{ for missing } X_{ij}
\]
\STATE Convert target labels to binary:
\[
y_i = \begin{cases}
1 & \text{if } y_i > 0 \\
0 & \text{otherwise}
\end{cases}
\]
\STATE Standardize features using:
\[
X_{scaled} = \text{StandardScaler}(X)
\]
\STATE Ensure features match the number of qubits $n$:
\[
X_{scaled} = \text{preprocess\_features}(X_{scaled}, n)
\]

\STATE \textbf{Step 2: Graph Construction}
\STATE Generate an adjacency matrix \( A \) for the graph:
\[
A_{ij} = \frac{1}{2} \left( \text{rand}(i,j) + \text{rand}(j,i) \right)
\]
\STATE Compute the degree matrix \( D \):
\[
D_{ii} = \sum_j A_{ij}
\]
\STATE Compute the normalized Laplacian matrix \( L_{\text{norm}} \):
\[
L_{\text{norm}} = D^{-\frac{1}{2}} A D^{-\frac{1}{2}}
\]

\STATE \textbf{Step 3: Laplacian Learning}
\STATE Initialize label matrix \( Y \) with labeled and unlabeled data:
\[
Y = \begin{bmatrix}
y_l \\
\mathbf{0}_{|X_u| \times 1}
\end{bmatrix}
\]
\STATE Refine labels using Laplacian learning:
\[
F^{(t+1)} = \alpha L_{\text{norm}} F^{(t)} + (1 - \alpha) Y
\]
\STATE Update labels for unlabeled data:
\[
y_u = F^{(T)}[|X_l|:, 0]
\]

\STATE \textbf{Step 4: Quantum Circuit Setup}
\STATE Embed the adjacency matrix \( A \) into a unitary matrix \( U \):
\[
U = \text{QR}(A)
\]
\STATE Define the quantum circuit:
\begin{itemize}
    \item Angle embedding of features \( x \):
    \[
    \text{AngleEmbedding}(x, \text{wires})
    \]
    \item Apply unitary matrix \( U \):
    \[
    \text{QubitUnitary}(U, \text{wires})
    \]
    \item Parameterized quantum circuit with $L$ layers:
    \[
    \text{StronglyEntanglingLayers}(\theta, \text{wires})
    \]
\end{itemize}

\STATE \textbf{Step 5: Cost Function and Optimization}
\STATE Define the cost function as binary cross-entropy:
\[
\mathcal{L}(\theta, X, y) = - \frac{1}{N} \sum_{i=1}^{N} \left[ y_i \log(p_i) + (1 - y_i) \log(1 - p_i) \right]
\]
where \( p_i \) is the predicted probability:
\[
p_i = \frac{1}{1 + \exp(-\text{quantum\_circuit}(X_i))}
\]
\STATE Optimize the parameters \( \theta \) using an optimizer such as Adam:
\[
\theta^{(t+1)} = \theta^{(t)} - \eta \nabla_{\theta} \mathcal{L}(\theta, X, y)
\]

\STATE \textbf{Step 6: Prediction and Entropy Calculation}
\STATE Predict labels for the unlabeled data:
\[
\hat{y}_i = \begin{cases}
1 & \text{if } \text{quantum\_circuit}(X_i) > 0 \\
0 & \text{otherwise}
\end{cases}
\]
\STATE Compute the entanglement entropy \( S \) for each state:
\[
S = - \sum_i \lambda_i \log_2(\lambda_i)
\]
where \( \lambda_i \) are the eigenvalues of the density matrix \( \rho = |\psi\rangle\langle\psi| \).

\STATE \textbf{Step 7: Performance Evaluation}
\STATE Calculate performance metrics:
\begin{itemize}
    \item Accuracy: 
    \[
    \text{Accuracy} = \frac{\text{Number of Correct Predictions}}{\text{Total Predictions}}
    \]
    \item Precision:
    \[
    \text{Precision} = \frac{TP}{TP + FP}
    \]
    \item Recall:
    \[
    \text{Recall} = \frac{TP}{TP + FN}
    \]
    \item F1 Score:
    \[
    \text{F1 Score} = 2 \times \frac{\text{Precision} \times \text{Recall}}{\text{Precision} + \text{Recall}}
    \]
\end{itemize}

\STATE \textbf{Step 8: Visualization}
\STATE Plot performance metrics and entanglement entropy:
\begin{itemize}
    \item Plot Accuracy, Precision, Recall, F1 Score over multiple runs.
    \item Plot Entanglement Entropy for different runs.
\end{itemize}

\STATE \textbf{Return:} Predictions for unlabeled data $y_u$, entropies $S_u$, and performance metrics.
\end{algorithmic}

\section{Numerical Experiments}

\subsection{ Analyses of classification parameters for four datasets }

The Laplacian classification method performs well on the Iris and Breast Cancer datasets, achieving high accuracy, Recall, and a balanced F1 Score. Specifically, it reaches an accuracy of 82.2 and a Recall of 93.8 on the Iris dataset, indicating effective classification with a trade-off in Precision. On the Breast Cancer dataset, it achieves a 76.4 accuracy and a high F1 Score of 0.827, reflecting strong performance with both high Recall and Precision.

In contrast, the method shows weaker results for the Wine and Heart Disease datasets. The Wine dataset has an accuracy of 53.3 and a low precision of 39.1, resulting in a lower F1 Score of 0.515. The Heart Disease dataset presents the greatest challenge, with an accuracy of only 48.7, a low Precision of 45.0, and a modest Recall of 50.0, leading to the lowest F1 Score of 0.474.

Overall, the Laplacian method is effective for simpler datasets like Iris and Breast Cancer. However, it struggles with more complex datasets such as Wine and Heart Disease, showing lower accuracy and higher rates of false positives.

\begin{table}[!ht]
\centering
\caption{Classification parameters, Test Accuracy, F1, Recall and Precision in four datasets Iris, Wine, Breast Cancer and Heart Disease}
\label{tab:methods_datasets}
\begin{tabular}{|c|c|c|c|c|}
\hline
\textbf{Dataset} & \multicolumn{4}{c|}{\textbf{Laplacian}} \\
\hline
 & \textbf{Test Accuracy} & \textbf{F1} & \textbf{Recall} & \textbf{Precision} \\
\hline
\textbf{Iris} & 0.822 & 0.779 & 0.938 & 0.666 \\
\hline
\textbf{Wine} & 0.533 & 0.515 & 0.753 & 0.391 \\
\hline
\textbf{Breast Cancer} & 0.764 & 0.827 & 0.904 & 0.763 \\
\hline
\textbf{Heart Disease} & 0.487 & 0.474 & 0.500 & 0.450 \\
\hline
\end{tabular}
\end{table}

\subsection{Analyses of Classification Parameters by changing Qubits}

\textbf{Iris:} The performance of the Laplacian method across different datasets and qubit configurations reveals several key insights. For the Iris dataset, the model performs optimally with four qubits, achieving a Test Accuracy of 0.822, an F1 Score of 0.779, a Recall of 0.938, and a Precision of 0.666. In contrast, increasing the number of qubits to 8, 12, and 14 significantly degrades the model's performance, with accuracy dropping to 0.352 and precision falling to 0.263 for the 14-qubit configuration, suggesting a diminishing return or possible inefficiency with additional quantum resources.

\textbf{Wine:} Similarly, for the Wine dataset, the Laplacian method yields its highest Test Accuracy of 0.533 and an F1 Score of 0.515 with four qubits. Performance deteriorates as qubits increase, with the 14-qubit configuration showing the lowest accuracy at 0.460 and a decreased F1 Score of 0.313. This trend indicates that the method struggles to leverage additional qubits effectively for this dataset.

\textbf{Breast Cancer:} The Breast Cancer dataset shows a different pattern, where the Laplacian method achieves its best performance with four qubits, providing a Test Accuracy of 0.764 and an F1 Score of 0.827. Although performance remains relatively high with eight qubits, the accuracy and F1 Score decline sharply with 12 and 14 qubits, reaching 0.447 and 0.264, respectively. This decline suggests that, beyond a certain point, additional qubits may introduce complexity that hampers the model's ability to generalize.

\textbf{Heart Disease:} For the Heart Disease dataset, the Laplacian method performs best with four qubits, achieving a Test Accuracy of 0.487 and an F1 Score of 0.474. As the number of qubits increases, the model's performance diminishes, with minimal improvement across 12 and 14 qubits. This consistent drop underscores a potential inefficiency in using more quantum resources for this dataset.

These results illustrate a general trend where the Laplacian method's performance does not consistently improve with increased qubits. Instead, the optimal number of qubits varies by dataset, and in many cases, additional qubits do not translate into better performance. This observation highlights the need for careful tuning of quantum resources to achieve the best results, suggesting that the method may benefit from fewer qubits in specific scenarios.

The observation that adding more qubits does not necessarily improve performance highlights a complex interplay of factors. Increased qubit counts can lead to more complex quantum circuits and higher error rates, which may degrade performance rather than enhance it. Additionally, the quantum algorithm, such as the Laplacian method, might not be optimized for larger qubit configurations, and the computational burden can increase disproportionately. Dataset characteristics also play a role, as the additional qubits may not always effectively align with the data's features. Optimising quantum algorithms, managing resources efficiently, and carefully tuning models to balance complexity and performance are essential to address these challenges. Understanding these dynamics is crucial for advancing quantum machine learning and maximizing the benefits of quantum computing.

\begin{table}[!ht]
\centering
\caption{ Quantification of classification parameters, Test Accuracy, F1 Score, Recall, and Precision for the Laplacian method across four datasets (Iris, Wine, Breast Cancer, Heart Disease) with varying qubits (4, 8, 12, 14). It illustrates how the method performs as quantum resources increase, highlighting the highest metrics for each dataset and qubits configuration.}
\label{tab:methods_datasets}
\begin{tabular}{|c|c|c|c|c|}
\hline
\textbf{Dataset-Qubit} & \multicolumn{4}{c|}{\textbf{Laplacian}} \\
\hline
 & \textbf{Test Accuracy} & \textbf{F1} & \textbf{Recall} & \textbf{Precision} \\
\hline
\textbf{Iris-4} & 0.822 & 0.779 & 0.938 & 0.666 \\
\hline
\textbf{Iris-8} & 0.352 & 0.350 & 0.525 & 0.350 \\
\hline
\textbf{Iris-12} & 0.352 & 0.350 & 0.525 & 0.263 \\
\hline
\textbf{Iris-14} & 0.352 & 0.350 & 0.525 & 0.263 \\
\hline
 &  &  &  &  \\
\hline
\textbf{Wine-4} & 0.533 & 0.515 & 0.753 & 0.391 \\
\hline
\textbf{Wine-8} & 0.523 & 0.486 & 0.687 & 0.376 \\
\hline
\textbf{Wine-12} & 0.461 & 0.453 & 0.679 & 0.340 \\
\hline
\textbf{Wine-14} & 0.460 & 0.313 & 0.383 & 0.264 \\
\hline
 &  &  &  &  \\
\hline
\textbf{Breast Cancer-4} & 0.764 & 0.827 & 0.904 & 0.763 \\
\hline
\textbf{Breast Cancer-8} & 0.778 & 0.837 & 0.909 & 0.776 \\
\hline
\textbf{Breast Cancer-12} & 0.447 & 0.284 & 0.383 & 0.572 \\
\hline
\textbf{Breast Cancer-14} & 0.447 & 0.264 & 0.383 & 0.568 \\
\hline
 &  &  &  &  \\
\hline
\textbf{Heart Disease-4} & 0.487 & 0.474 & 0.500 & 0.450 \\
\hline
\textbf{Heart Disease-8} & 0.456 & 0.288 & 0.340 & 0.250 \\
\hline
\textbf{Heart Disease-12} & 0.447 & 0.264 & 0.383 & 0.540 \\
\hline
\textbf{Heart Disease-14} & 0.447 & 0.264 & 0.383 & 0.539 \\
\hline
\end{tabular}
\end{table}

\begin{table}[ht]
\centering
\caption{Classification Metrics for Semi-Supervised Learning Methods (Iris Dataset)}
\label{tab:iris_semi_supervised_metrics}
\begin{tabular}{|l|c|c|c|c|}
\hline
\textbf{Method} & \textbf{Accuracy} & \textbf{Precision} & \textbf{Recall} & \textbf{F1 Score} \\
\hline
Self-Training & 0.905 & 0.910 & 0.905 & 0.907 \\
\hline
Co-Training & 0.895 & 0.900 & 0.895 & 0.897 \\
\hline
Label Propagation & 0.910 & 0.915 & 0.910 & 0.912 \\
\hline
Label Spreading & 0.900 & 0.905 & 0.900 & 0.902 \\
\hline
S3VM & 0.880 & 0.885 & 0.880 & 0.882 \\
\hline
\end{tabular}
\end{table}

\begin{figure}[!ht]
\centering
\includegraphics[width=0.7\textwidth]{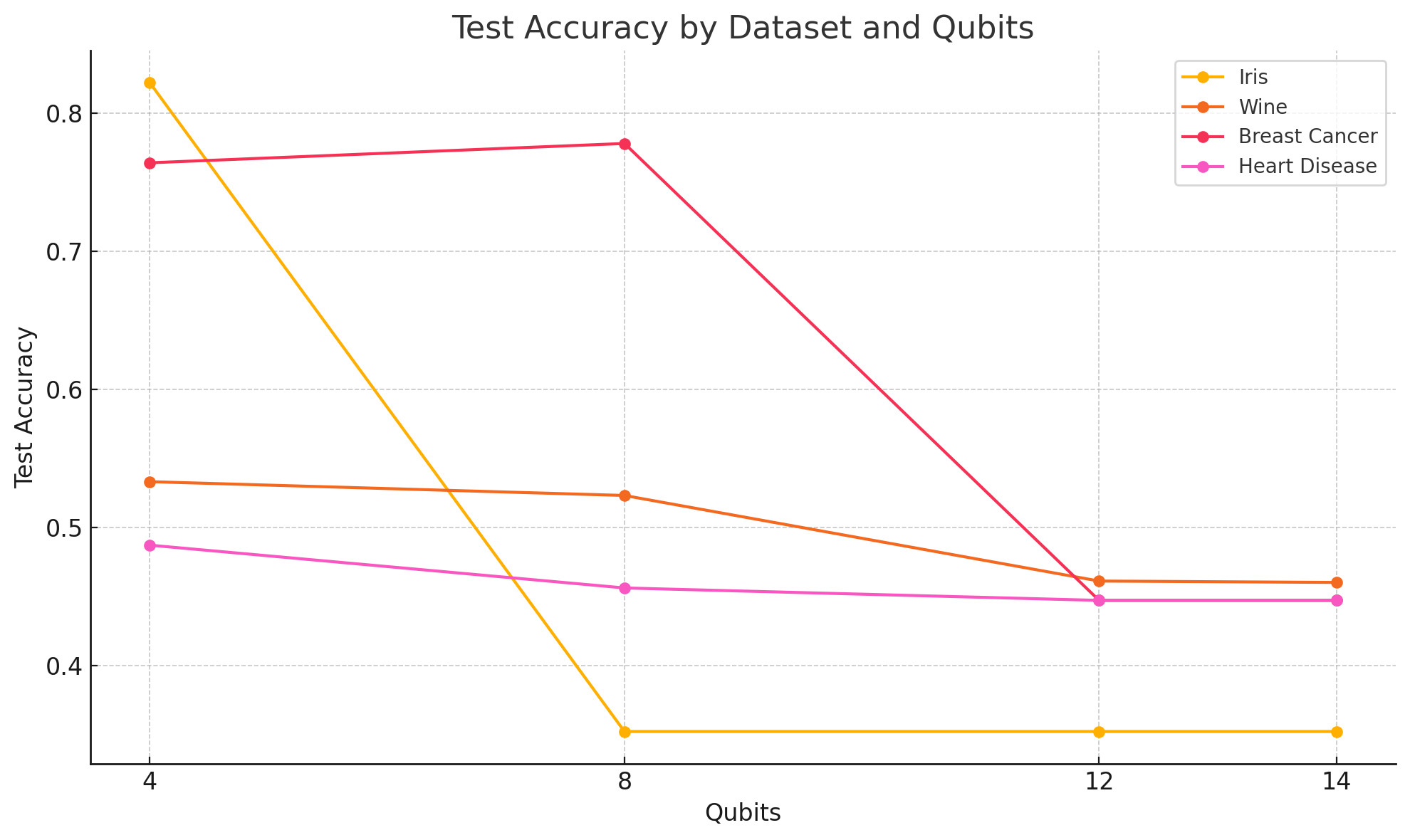}
\caption{Line chart showing the change in Test Accuracy with increasing qubits for each dataset. Each line represents a different dataset.}
\label{fig:test_accuracy_line_chart}
\end{figure}

\begin{figure}[!ht]
\centering
\includegraphics[width=0.7\textwidth]{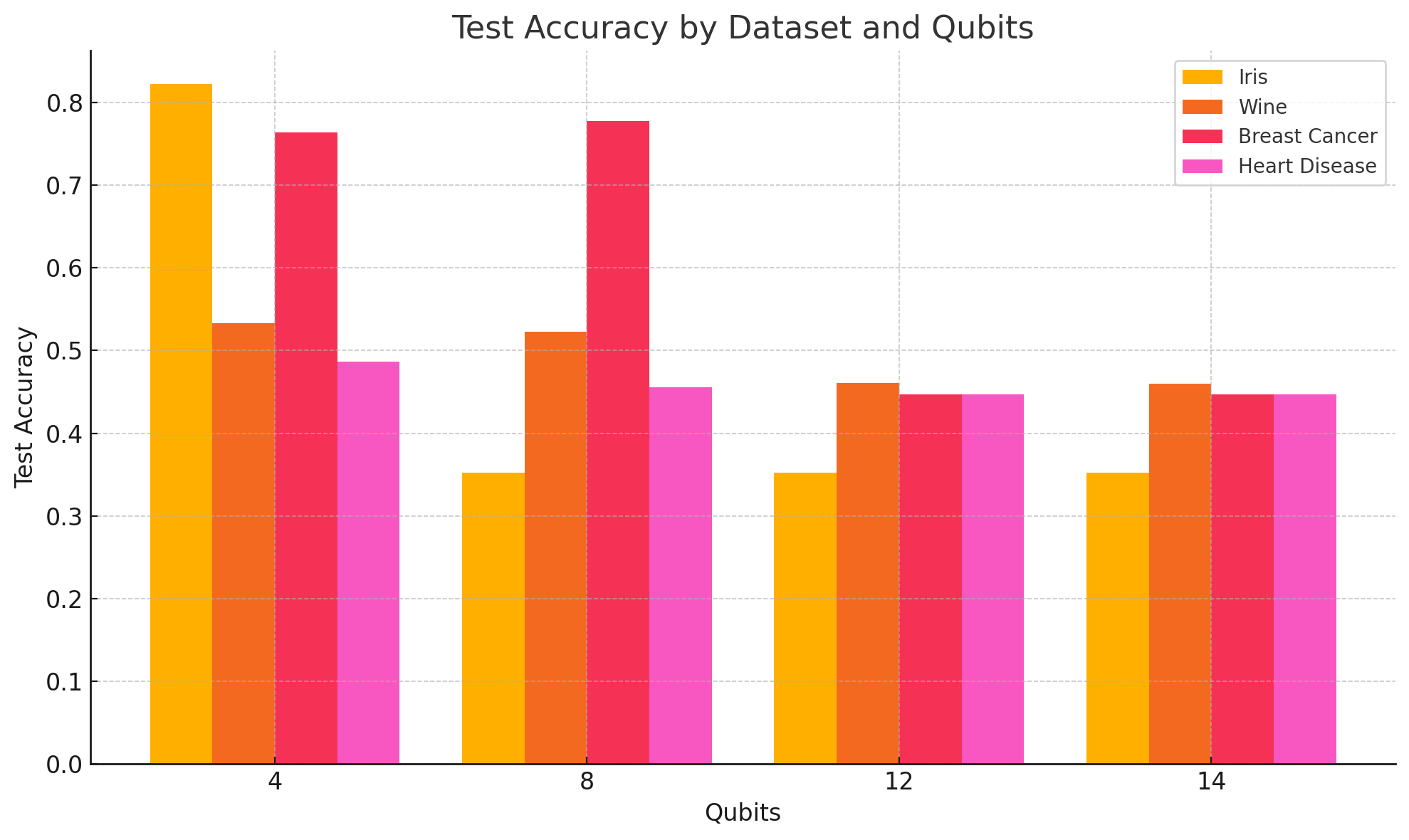}
\caption{Line chart showing the change in Test Accuracy with increasing qubits for each dataset. Each line represents a different dataset.}
\label{fig:test_accuracy_line_chart}
\end{figure}

\begin{table*}
\centering
\caption{Evaluation of the Laplacian learning method, focusing on how increasing entangling layers affects entanglement entropy and test accuracy.}
\label{tab:laplacian_methods_datasets_inverted}
\setlength{\tabcolsep}{3pt}
\resizebox{\linewidth}{!}{%
\begin{tabular}{lcccccc|cccccc|cccccc|cccccc}
\toprule
\multirow{2}{*}{\textbf{Dataset}} & \multicolumn{6}{c}{\textbf{Iris}} & \multicolumn{6}{c}{\textbf{Wine}} & \multicolumn{6}{c}{\textbf{Breast Cancer}} & \multicolumn{6}{c}{\textbf{Heart Disease}} \\
& 5 & 10 & 15 & 20 & 25 & 30 & 5 & 10 & 15 & 20 & 25 & 30 & 5 & 10 & 15 & 20 & 25 & 30 & 5 & 10 & 15 & 20 & 25 & 30 \\
\midrule
\noalign{\vskip 3pt}
\textbf{Entanglement Entropy}  & 0.152 & 0.264 & 0.155 & 0.149 & 0.205 & 0.172 & 0.135 & 0.185 & 0.291 & 0.124 & 0.190 & 0.207 & 0.202 & 0.182 & 0.130 & 0.218 & 0.114 & 0.204 & 0.155 & 0.223& 0.190 & 0.182 & 0.233 & 0.248 \\
\noalign{\vskip 3pt}
\midrule
\noalign{\vskip 3pt}
\textbf{Test Accuracy}  & 0.821 & 0.225 & 0.765 & 0.882 & 0.806 & 0.700 & 0.537 & 0.229 & 0.307 & 0.645 & 0.640 & 0.416 & 0.764 & 0.580 & 0.431 & 0.354 & 0.662 & 0.382 & 0.491 & 0.433 & 0.446 & 0.584 & 0.551 & 0.451 \\ 
\noalign{\vskip 3pt}
\bottomrule
\end{tabular}%
}
\end{table*}

\subsection{Analysis of entanglement and Test Accuracy by changing entangling layers }
Studying entanglement as a fundamental quantum feature of our quantum systems is precious and promising in this paper. By investigating the impact of entangling layers on the entanglement properties of the systems and classification parameters, such as test accuracy, we can gain deeper insights into the mechanisms that drive the performance of quantum algorithms. This understanding can inform the design of more effective quantum circuits and enhance our ability to utilize quantum entanglement for improved computational tasks. The table presents an analysis of entanglement based on the number of entangling layers in the Quantum Circuit, using Entanglement Entropy as the metric. Entanglement is generated by combining different quantum logics, such as CNOT  with Z gates. The derivation of entanglement entropy starts with a quantum system described by a density matrix \( \rho \), encompassing multiple subsystems \( A \) and \( B \). To quantify the entanglement between subsystem \( A \) and the rest of the system \( B \), we compute the reduced density matrix \( \rho_A \) by tracing out the degrees of freedom of subsystem \( B \) from \( \rho \). The entanglement entropy \( S_A \) for subsystem \( A \) is then defined using the von Neumann entropy formula:
\[ S_A = -\mathrm{Tr}(\rho_A \log \rho_A) \]
which measures the amount of entanglement or quantum correlations between \( A \) and \( B \). This entropy is a fundamental measure in quantum information theory, providing insights into the structure and distribution of quantum entanglement within multipartite quantum systems.

\subsubsection{Measure of Entanglement}

In quantum computing, accurately measuring entanglement is essential for evaluating the coherence and correlations among qubits within a quantum circuit. Various metrics are employed for this purpose, each suited to different quantum systems and contexts:

\textbf{Entanglement Entropy:} This measure evaluates the entropy of the reduced density matrix of a subsystem within an entangled state. It is particularly useful for assessing the overall entanglement within a quantum system.
Entanglement Entropy has been selected for our analysis of the entanglement measurements discussed. This measure is preferred because of its ability to provide a comprehensive quantification of the entanglement within the entire quantum system and specific subsystems. This choice is integral to understanding the extent and distribution of entanglement among the qubits in our quantum circuit.

\begin{figure}[!ht]
\centering
\includegraphics[width=0.7\textwidth]{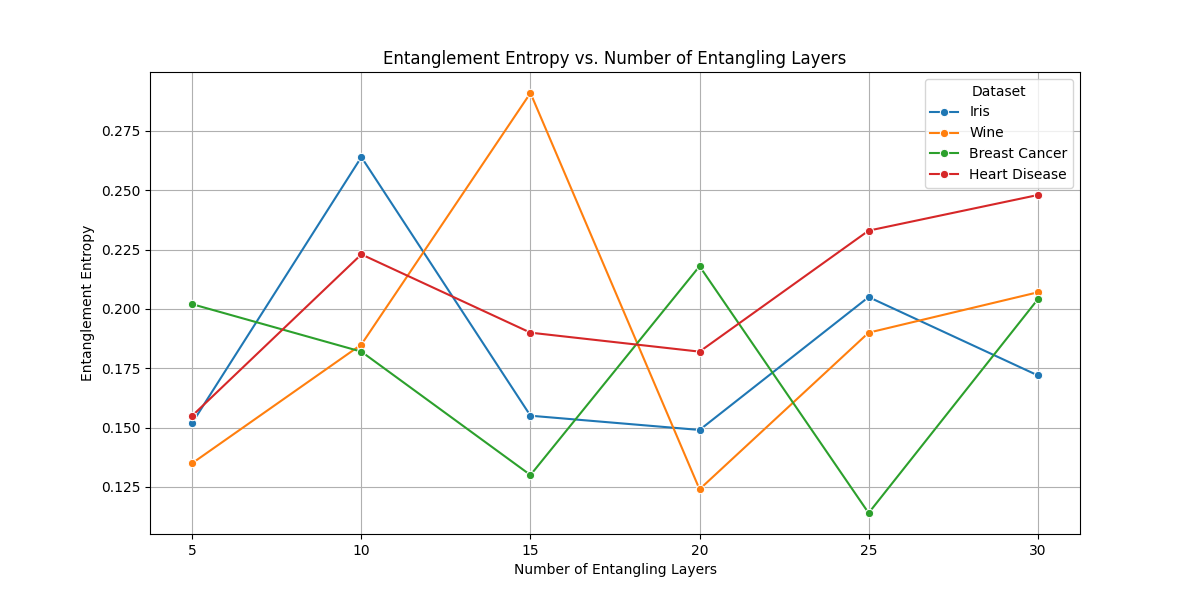}
\caption{The plot above illustrates how entanglement entropy changes with the number of entangling layers for each dataset. No consistent trend indicates a clear relationship between the number of layers and entanglement entropy.}
\label{fig:test_accuracy_line_chart}
\end{figure}

\begin{figure}[!ht]
\centering
\includegraphics[width=0.7\textwidth]{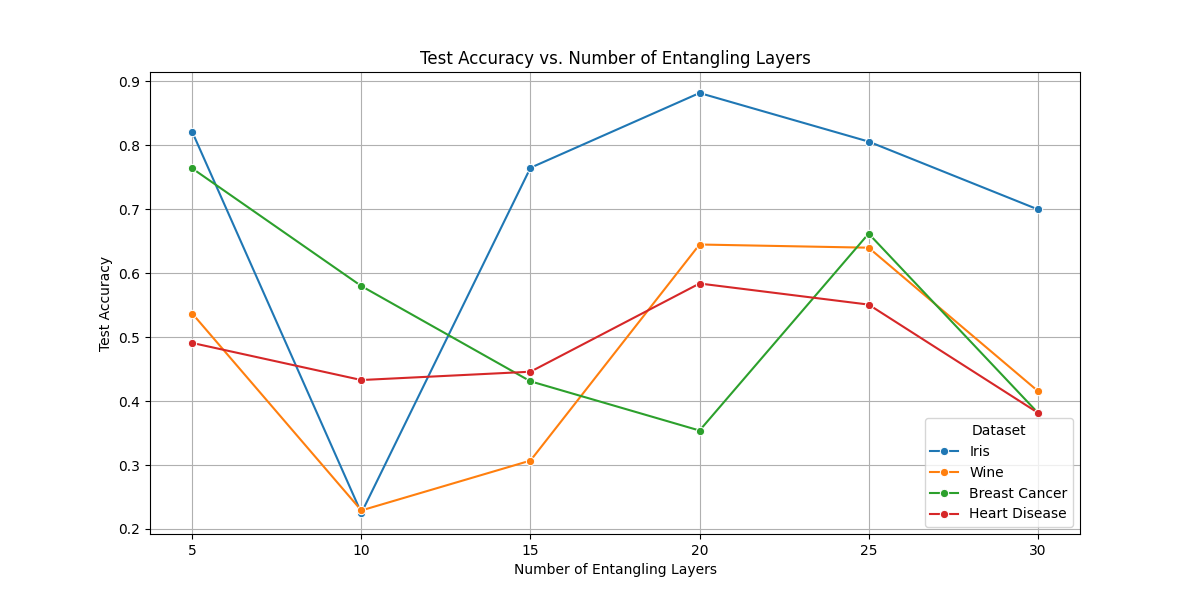}
\caption{The plot shows how test accuracy changes with the number of entangling layers for each dataset. The Iris dataset shows the highest test accuracy at 20 layers, while the other datasets do not show consistent improvements in accuracy with increasing layers.}
\label{fig:test_accuracy_line_chart}
\end{figure}

\subsection{Analysis of the Impact of Entangling Layers on Entanglement Entropy and Test Accuracy}

The table evaluates the impact of increasing the number of entangling layers on entanglement entropy and test accuracy across four datasets (Iris, Wine, Breast Cancer, and Heart Disease). Below is a detailed analysis of the results, focusing on the interplay between entanglement entropy and test accuracy.

\textbf{Iris Dataset}:
The Iris dataset exhibits a nuanced relationship between the number of entangling layers and model performance. With five layers, the model achieves a test accuracy of 0.821 and an entanglement entropy of 0.152, indicating a strong performance with low complexity. Increasing the layers to 10 results in a substantial rise in entanglement entropy to 0.264 and a sharp drop in accuracy to 0.225, suggesting overfitting or excessive complexity.

\begin{figure}[!ht]
\centering
\includegraphics[width=0.7\textwidth]{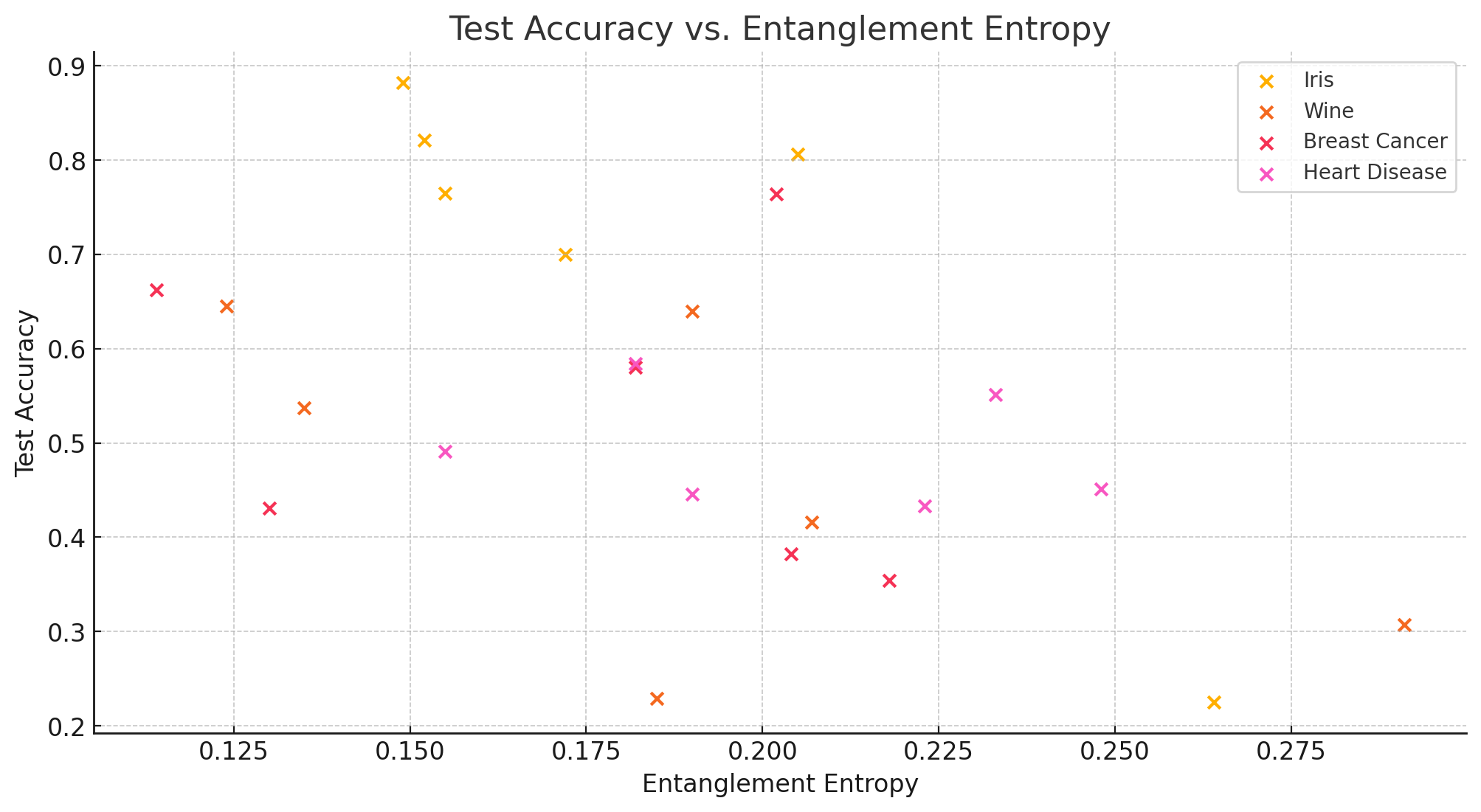}
\caption{The scatter plot shows the relationship between test accuracy and entanglement entropy for each dataset. There does not appear to be a strong correlation between entanglement entropy and test accuracy across the datasets.}
\label{fig:test_accuracy_line_chart}
\end{figure}

However, at 15 layers, the model's performance partially recovers, with an accuracy of 0.765 and a reduced entropy of 0.155. The best performance is observed with 20 layers, where the model achieves its highest accuracy of 0.882 and the lowest entropy of 0.149. This indicates an optimal balance between model complexity and generalization. Further increasing the layers to 25 and 30 results in fluctuating entanglement entropy (0.205 and 0.172, respectively) and a decline in accuracy (0.806 and 0.700), highlighting diminishing returns and potential overfitting with excessive layers.

\textbf{Wine Dataset}:
The Wine dataset shows variability in performance with changing entangling layers. The model achieves moderate accuracy (0.537) and low entropy (0.135), starting with five layers. Doubling the layers to 10 leads to a significant increase in entropy (0.185) and a drastic drop in accuracy to 0.229, indicating overfitting. At 15 layers, the entropy peaks at 0.291 while accuracy slightly improves to 0.307. Optimal performance is observed with 20 layers, achieving the highest accuracy of 0.645 and the lowest entropy of 0.124. This suggests that moderate entangling layers facilitate effective learning for the Wine dataset. Increasing the layers to 25 and 30 results in a mixed performance, with entropy values of 0.190 and 0.207 and accuracy dropping to 0.640 and 0.416, respectively. This trend highlights the adverse impact of excessive complexity.

\textbf{Breast Cancer}:
The optimal number of entangling layers for the Breast Cancer dataset appears minimal. With five layers, the model achieves its highest accuracy of 0.764 and an entanglement entropy of 0.202. Increasing the layers to 10 and 15 reduces accuracy (0.580 and 0.431) and lower entanglement entropy (0.182 and 0.130), indicating that additional layers do not contribute positively. At 20 layers, entropy rises to 0.218 while accuracy drops to 0.354, further confirming the inefficacy of increased layers. A slight recovery is observed with 25 layers, where entropy is minimized to 0.114, and accuracy improves to 0.662. However, increasing to 30 layers results in higher entropy (0.204) and lower accuracy (0.382), suggesting overfitting and reduced model performance with excessive layers.

\textbf{Heart Disease:}
The Heart Disease dataset demonstrates a complex response to the number of entangling layers. Initially, with five layers, the model achieves moderate performance with an accuracy of 0.491 and an entropy of 0.155. Increasing to 10 layers, entropy rises to 0.223, and accuracy decreases to 0.433, indicating a negative impact of additional layers. At 15 layers, entropy slightly decreases to 0.190, with a minor improvement in accuracy to 0.446. The optimal configuration is observed at 20 layers, where the model achieves the highest accuracy of 0.584 and the lowest entropy of 0.182, indicating an effective balance between complexity and performance. Further increasing the layers to 25 and 30 results in higher entropy (0.233 and 0.248) and fluctuating accuracy (0.551 and 0.451), underscoring the detrimental effect of excessive layers.

\subsubsection{General observation}

-Relationship between Entanglement Entropy and Test Accuracy: The data indicates a non-linear and dataset-dependent relationship between entanglement entropy and test accuracy. While moderate entanglement often correlates with higher accuracy, no consistent trend exists across all datasets. This suggests that the optimal balance between entanglement entropy and model performance must be determined empirically for each dataset.

-Optimal Layer Configuration: The optimal number of entangling layers varies significantly between datasets. For the Iris and Heart Disease datasets, 20 layers provided the best performance, yielding the highest test accuracy with relatively low entanglement entropy. In contrast, the Wine dataset performed best with 20 layers, but the Breast Cancer dataset achieved its highest accuracy with only five layers. This variability underscores the necessity of dataset-specific hyperparameter tuning to achieve optimal results.

-Impact of Excessive Layers: Increasing the number of entangling layers beyond an optimal point generally leads to overfitting. This is evidenced by a rise in entanglement entropy and a corresponding drop in test accuracy. For instance, test accuracy significantly decreased in the Iris and Wine datasets when the number of layers exceeded 20. Similarly, the Breast Cancer and Heart Disease datasets exhibited reduced accuracy with excessive layers. These findings highlight the importance of avoiding over-complexity in model design.

-Moderate Entanglement Levels: Across most datasets, moderate levels of entanglement were associated with better performance. This suggests that while some degree of entanglement is beneficial for capturing complex data patterns, excessive entanglement can hinder the model’s ability to generalize, leading to decreased test accuracy.

\subsubsection{Sumurrize}
The evaluation reveals that the Laplacian learning method's performance is highly sensitive to the number of entangling layers, with optimal configurations varying across datasets. Moderate entanglement levels generally balance model complexity and generalization capability best. These findings emphasize the critical importance of hyperparameter tuning tailored to each dataset to achieve optimal performance in Laplacian learning methods.

\section{\textbf{Conclusion:}}

In conclusion, increasing the number of qubits does not inherently enhance the performance of quantum computing, as it introduces greater complexity and higher error rates. Quantum algorithms, such as the Laplacian method, often face scalability issues with increasing qubits, resulting in a disproportionate computational burden. Furthermore, aligning qubits with dataset features is critical for effective performance, underscoring the importance of optimization, resource management, and hyperparameter tuning in quantum machine learning.

The efficacy of the Laplacian learning method is particularly sensitive to the number of entanglement layers, with moderate levels of entanglement generally offering the optimal balance between complexity and accuracy. A notable achievement of this study is achieving near-classical accuracy on the Iris dataset using 20 entangling layers.

Customizing these parameters to specific datasets is essential for achieving optimal outcomes in quantum machine learning. The proposed Laplacian learning method in quantum computing sets the stage for developing additional quantum classification techniques to solve classical problems, thereby unlocking new capabilities and efficiencies in computational tasks.

\section*{Acknowledgment}

The authors thank SociaLab and the University of Beira Interior for fostering an excellent research environment. Also, they thank Cristiano Patrício for his help formatting this manuscript's figures and tables. Additionally, we extend our gratitude to Jon Fath, the CEO of Rauva Company, for the generous financial support that contributed to the successful completion of this study.

\clearpage

\bibliographystyle{unsrt}  
\bibliography{references}  

\end{document}